\documentclass[letterpaper,twocolumn,10pt]{article}
\usepackage{usenix}

\everypar{\looseness=-1}
\usepackage{todonotes}
\usepackage{multirow}
\usepackage{tabls}
\usepackage{wrapfig}
\usepackage{subfigure}
\usepackage{amsmath,amsthm,amssymb}
\usepackage{etoolbox}
\usepackage{array}
\usepackage{tikz}
\usepackage{pgfplots}
\usetikzlibrary{shapes,arrows,arrows.meta, positioning,shapes.misc,decorations.markings,pgfplots.groupplots}
\usepackage{adjustbox}
\usepackage{hyperref}
\usepackage{booktabs}
\usepackage{lipsum,multicol}
\usepackage[numbers]{natbib}
\usepackage{xurl}

\usepackage{mdframed}
\newcommand{\sectioncolor}{violet}

\usepackage{etoolbox}

\makeatletter
\patchcmd{\@maketitle}{\raggedright}{\centering}{}{}
\patchcmd{\@maketitle}{\raggedright}{\centering}{}{}
\makeatother

\begin{document}

\title{MOTIF: A Large Malware Reference Dataset with Ground Truth Family Labels}

\author{
Robert J. Joyce,\textsuperscript{\rm 1,\rm 2}
Dev Amlani,\textsuperscript{\rm 2,\rm 3}
Charles Nicholas,\textsuperscript{\rm 2}
Edward Raff,\textsuperscript{\rm 1,\rm 2}\\
\textsuperscript{\rm 1}Booz Allen Hamilton,
\textsuperscript{\rm 2}University of Maryland, Baltimore County,
\textsuperscript{\rm 3}Marriotts Ridge High School\\
joyce\_robert2@bah.com, amlanid35@gmail.com, nicholas@umbc.edu, raff\_edward@bah.com
}

\maketitle

\begin{abstract}
Malware family classification is a significant issue with public safety and research implications that has been hindered by the high cost of expert labels. The vast majority of corpora use noisy labeling approaches that obstruct definitive quantification of results and study of deeper interactions. In order to provide the data needed to advance further, we have created the Malware Open-source Threat Intelligence Family (MOTIF) dataset. MOTIF contains 3,095 malware samples from 454 families, making it the largest and most diverse public malware dataset with ground truth family labels to date, nearly $3\times$ larger than any prior expert-labeled corpus and $36\times$ larger than the prior Windows malware corpus. MOTIF also comes with a mapping from malware samples to threat reports published by reputable industry sources, which both validates the labels and opens new research opportunities in connecting opaque malware samples to human-readable descriptions. This enables important evaluations that are normally infeasible due to non-standardized reporting in industry. For example, we provide aliases of the different names used to describe the same malware family, allowing us to benchmark for the first time accuracy of existing tools when names are obtained from differing sources. Evaluation results obtained using the MOTIF dataset indicate that existing tasks have significant room for improvement, with accuracy of antivirus majority voting measured at only 62.10\% and the well-known AVClass \cite{avclass} tool having just 46.78\% accuracy. Our findings indicate that malware family classification suffers a type of labeling noise unlike that studied in most ML literature, due to the large open set of classes that may not be known from the sample under consideration \cite{Northcutt2021,10.1145/3351095.3372862,NEURIPS2020_c6102b37,Patrini2017,Liu2016,Nicholson2015,Frenay2014,Natarajan2013}.

\end{abstract}

\section{Introduction}
\label{sec:introduction}

A malware family is a collection of malicious files that are derived from a common source code. Classifying a malware sample into a known family provides valuable insights about the behaviors it likely performs and can greatly aid triage and remediation efforts \cite{agtr}. Using manual analysis to label a large quantity of malware is generally considered to be intractable, and developing methods to automatically classify malware with high fidelity has long been an area of interest. Approximate labeling methods are typically employed due to the difficulty of obtaining family labels with ground truth confidence, often using a single antivirus or the majority vote of a collection of them. However, it is also common for malware family classifiers to make their classification decisions based upon antivirus signatures \cite{avclass} or to train them using antivirus-based labels \cite{huang}. This has the potential to bias evaluation results. The difficulty of constructing datasets and evaluating them in a manner that avoids over-estimating accuracy has been a chronic problem for over two decades \cite{Rossow2012,235493,Jordaney2016,Christodorescu:2004:TMD:1007512.1007518,Marx2000}. 

In order to improve current evaluation practices, especially evaluation of antivirus-based malware classifiers, we introduce the Malware Open-source Threat Intelligence Family datset (MOTIF). Containing 3,095 samples from 454 families, the MOTIF dataset is the largest and most diverse public malware dataset with expert-derived, ground truth confidence family labels to date. MOTIF is the first dataset to provide a comprehensive alias mapping that includes aliases derived from both open-source reporting and antivirus signatures. Each malware sample is associated with a report written by a domain expert, which acts as the high confidence source for the label. This provides a new direction to explore the intersection of document processing and natural language understanding with malware family detection, as well as the labeling and documentation needed to design experiments with higher confidence then previously possible. 

For the remainder of this section we provide an overview of the methods historically used to label malware and perform a survey of notable reference datasets.
In Section \ref{sec:motif} we describe the contents of the MOTIF dataset and the methodology used to construct it. In Section \ref{sec:experiments} we use MOTIF to benchmark the performance of a well-known malware labeling tool, four hash-based clustering algorithms, and two ML classifiers. Further discussion and conclusions are offered in Section \ref{sec:conclusion}.

\subsection{Malware Dataset Labeling Strategies}

\textbf{Manual Labeling.} Malware family labels obtained via manual analysis are said to have ground truth confidence. Although manual analysis is not perfectly accurate, the error rate is considered to be negligible \cite{mohaisen2015}. Unfortunately, manual analysis is extremely time consuming. It can take an average of ten hours for a human analyst to fully analyze a previously unseen malware sample \cite{mohaisen2013}. Although a full analysis may not be necessary to determine the malware family, the degree of difficulty and cost that manual labeling imposes is evident. Manually labeling a large quantity of malware quickly becomes intractable, and we are not aware of any reference datasets in which the creators manually analyzed all malware samples themselves in order to obtain ground truth family labels.

\textbf{Open-Source Threat Intelligence Reports.}
Hundreds of open-source threat intelligence reports containing detailed analysis of malware samples are published by reputable cybersecurity organizations each year, mainly in the form of blog posts and whitepapers. These reports often focus on a specific family, campaign, or threat actor, and include the evidence the analyst uses to reach their conclusion -- giving high confidence of correct labeling. In addition to analysis of the related malware samples, they frequently include file hashes and other indicators of compromise (IOCs) related to the cyberattack. If the family labels published in a report were obtained using manual analysis, then we say they have ground truth confidence. The MOTIF dataset was constructed by processing thousands of these reports and aggregating labeled file hashes from them. Although this is a more scalable method of obtaining ground truth family labels, it cannot be automated and it is restricted to the content published by the aforementioned organizations. This is due to the high cost of manual analysis, with expert practitioners reporting days-to-weeks of effort to reverse a single file \cite{Votipka2019}.

\textbf{Approximate Labeling Methods.} The remaining methods for obtaining malware reference labels are by far the most commonly employed, but they do not provide ground truth confidence. Cluster labeling is an approach in which a dataset is first clustered and then one malware sample per cluster is manually analyzed. The label for that sample is applied to the entire cluster \cite{malsign, nappa}. However, this approach still requires some manual analysis and it relies on the precision of the selected clustering algorithm, which is often custom-made and not rigorously evaluated. A common, automated method for labeling malware is scanning with a single antivirus. However, antivirus signatures are frequently incorrect and can lack family information entirely \cite{botacin, mohaisen2014}. Furthermore, antivirus vendors often use different names for the same malware family, which we refer to as \emph{aliases}. Many malware families have one or more aliases, causing widespread labeling inconsistencies \cite{avclass}. Another typical approach to malware labeling is antivirus majority voting, in which a collection of antivirus products vote on the label. Although regarded as highly accurate, prior work has not quantified this assumption \cite{avclass}.
Antivirus majority voting may also cause a selection bias, because the samples for which a majority agree on a label are likely the ``easiest" to classify \cite{li}. AVClass, a leading public tool for performing automatic malware family classification, was recently used to label the EMBER2018 dataset. When provided an antivirus scan report for a malware sample, AVClass attempts to aggregate the many antivirus signatures in the report into a single family label. It does this by normalizing each antivirus signature, resolving aliases, and using plurality voting to predict the family. AVClass is open source, simple to use, and does not require the malware sample to obtain a label, making it a popular choice as a malware classifier since its release in 2016 \cite{avclass}. In Section \ref{sec:experiments} we compute the accuracy of both antivirus majority voting and AVClass for the first time and they correctly predict malware families in the MOTIF dataset just 62.10\% and 46.78\% of the time respectively.

\textbf{Non-Family Based Labeling.} Malware family labels are not perfect, and no oracle based categorization of family, or even if a program is a virus, is possible due to the halting problem ~\cite{Cohen1987}. Our definition of a malware family as being derived from common source code is within normal industry use\footnote{See \url{https://www.gdatasoftware.com/blog/malware-family-naming-hell} for an informative introduction to some of the challenges with malware family names}, but indeed the lack of objective means to group families leads some to desire different means of grouping. Alternative approaches to grouping malware exist, such as by functionality ~\cite{capa}. However, all approaches to malware grouping run into different versions of the same set of problems. Malware is written by an active adversary who attempts to evade and mislead analysts, including complex code obfuscations and misdirection via code theft to slow analysts and thwart automation~\cite{Botacin2020,Arp2020,Li2017,Giacinto2011}. Even standard countermeasures such as packing, which hides the original source code of a program from static analysis, are poorly understood and difficult to circumvent ~\cite{Aghakhani2020}.  Dynamic analysis of running malware is not full-proof for similar reasons. 
The process is extraordinarily expensive and time-consuming, and malware may have numerous means of detecting instrumentation, altering behavior, and only executing on target victims, if sufficiently motivated~\cite{Wampler2019,Egele2017}. The various pros and cons of alternative labeling schemes is beyond the scope of our study, and focuses on family labels given its ubiquity in academic research and industry practice. 

\begin{table}[h]
\centering
\caption{Public Datasets with Ground Truth Family Labels}
\label{tab:groundtruthdatasets}
\resizebox{\columnwidth}{!}{%
\begin{tabular}{@{}lrrlll@{}}
\toprule
Name & Samples & Families & Platform & Collection Period & Labeling Method\\
\midrule
    MOTIF (our work)  & 3,095 & 454 & Windows & Jan. 2016 - Jan. 2021 & Threat Reports\\
    MalGenome \cite{malgenome} & 1,260 & 49 & Android & Aug. 2010 - Oct. 2011 & Threat Reports\\
    Variant \cite{upchurch} & 85 & 8 & Windows & Jan. 2014 & Threat Reports\\
\bottomrule
\end{tabular}
}
\end{table}

\subsection{Notable Malware Reference Datasets}

Statistics about malware datasets with ground truth family labels are shown in Table \ref{tab:groundtruthdatasets}. Labels for all three datasets were gathered from open-source threat reports. MOTIF is the largest, and by far the most diverse of these datasets, and it contains the most modern malware samples. Variant contains just 85 malware samples from eight families, all apparently from a single report published by the DHS NCCIC in 2014 \cite{dhs_nccic}, with additional analysis provided in a second report by Dell Secureworks \cite{dell_secureworks}. MalGenome is nearly a decade old, as of the time of writing \cite{malgenome}.

\begin{table}[h]
\centering
\caption{Notable Public Datasets With Imperfect Labeling%
}
\resizebox{\columnwidth}{!}{%
\label{tab:publicdatasets}
\begin{tabular}{@{}lrrlll@{}}
\toprule
Name & Samples & Families & Platform & Collection Period & Labeling Method\\ \midrule
    Malheur \cite{malheur} & 3,133 & 24 & Windows & 2006 - 2009 & AV Majority Vote\\
    AMD \cite{AMD} & 24,553 & 71 & Android & 2010 - 2016 & Cluster Labeling\\
    Drebin \cite{drebin} & 5,560 & 179 & Android & Aug. 2010 - Oct. 2012 & AV Majority Vote\\
    VX Heaven \cite{vxheaven} & 271,092 & 137 & Windows & 2012 or earlier & Single AV\\
    Malicia \cite{malicia} & 11,363 & 55 & Windows & Mar. 2012 - Mar. 2013 & Cluster Labeling\\
    Kaggle \cite{kaggle} & 10,868 & 9 & Windows & Feb. 2015 or earlier & Susp. Single AV\\
    Malpedia \cite{malpedia} & 5,862 & 2,165 & Both & 2017 - ongoing & Hybrid \\
    MalDozer \cite{maldozer} & 20,089 & 32 & Android & Mar. 2018 or earlier & Susp. Single AV\\
    EMBER2018 \cite{ember} & 485,000 & 3,226 & Windows & 2018 or earlier & AVClass \\
    \bottomrule
\end{tabular}
}
\end{table}

Table \ref{tab:publicdatasets} displays public malware datasets with approximate family labels. The collection periods of the VX Heaven, Kaggle, and MalDozer datasets are undocumented; we use publication dates as an upper bound for the end of the collection period. The VX Heaven site operated between 1999 and 2012 and the dataset is considered to contain very outdated malware. The VX Heaven dataset was labeled using the Kaspersky antivirus \cite{vxheaven} and we suspect that the Kaggle dataset was labeled using Windows Defender \cite{windows_defender} due to the use of the label \texttt{Obfuscator.ACY} (a signature that Defender uses for obfuscated malware). The labeling method for MalDozer was not disclosed; however, the formatting of the provided family names suggests a single antivirus was used. The original EMBER dataset does not have family labels, but an additional 1,000,000 files (both malicious and benign) were released in 2018 \cite{loi2021}. 485,000 of these files are malware samples with AVClass labels. The Malpedia dataset is the most similar to MOTIF. Some of the labels in the Malpedia dataset were obtained using open-source reporting, and manual analysis was used for dumping and unpacking some malware samples. However, other family labels were derived using automated methods such as YARA rules and similarity analyses of unpacked files to known malware samples \cite{malpedia}. Therefore, we do not consider the Malpedia dataset to have full ground-truth confidence for all labels. 

\begin{table}
\centering
\caption{Notable Private  Datasets With Imperfect Labeling}
\label{tab:privatedatasets}
\resizebox{\columnwidth}{!}{%
\begin{tabular}{@{}lrrlll@{}}
\toprule
Name & Samples & Families & Platform & Collection Period & Labeling Method\\ \midrule
    Malsign \cite{malsign} & 142,513 & Unknown & Windows & 2012 - 2014 & Cluster labeling\\ 
    MaLabel \cite{mohaisen2015} & 115,157 & $\ge$ 80 & Windows & Apr. 2015 or earlier & AV Majority Vote\\
    MtNet \cite{huang} & 1,300,000 & 98 & Windows & Jun. 2016 or earlier & Hybrid\\ \bottomrule
\end{tabular}
}
\end{table}

The datasets in Table \ref{tab:privatedatasets} are not publicly available and some information about their contents is unknown. The majority of the files in the Malsign dataset are potentially unwanted applications (PUAs), rather than malware. Reference labels for Malsign were produced via clustering on statically-extracted features \cite{malsign}. Of the 115,157 samples in MaLabel, 46,157 belong to 11 large families and the remaining 69,000 samples belong to families with fewer than 1,000 samples each. The total number of families in the dataset is unknown \cite{mohaisen2015}. Microsoft provided the authors of the MtNet malware classifier with a dataset of 1.3 million malware samples, labeled using a combination of antivirus labeling and manual labeling. Due to the enormous scale of the dataset and the source that provided it, we suspect that the vast majority of the files were labeled using Windows Defender.

A lack of family diversity is apparent in many of the datasets we surveyed, most notably in the Variant and Kaggle datasets. The Windows datasets in particular tend to use approximate labeling methods, lack family diversity, and contain old malware. There is also a dearth of notable datasets containing malware that targets other platforms (e.g. Linux, macOS, and iOS), but this is work beyond our paper's scope.

\section{The MOTIF Dataset}
\label{sec:motif}

The MOTIF dataset was constructed by surveying all open-source threat intelligence reports published by 14 major cybersecurity organizations during a five-year period. During this process, we meticulously reviewed thousands of these reports to determine the ground truth families of 4,369 malware samples. 
The organizations that published these reports included only the file hashes, to avoid distributing live malware samples. Hence, much of the malware discussed in these reports was inaccessible to us, and MOTIF was limited to only the malware samples which we could obtain. Furthermore, although we originally intended the dataset to include Windows, macOS, and Linux malware, we found that the vast majority of malware samples in our survey targeted Windows. In order to standardize the dataset, we elected to discard all files not in the Windows Portable Executable (PE) file format. As a result, the MOTIF dataset includes disarmed versions of 3,095 PE files from the survey that were in our possession, labeled by malware family with ground truth confidence. All files were disarmed by replacing the values of their \texttt{OPTIONAL\_HEADER.Subsystem} and \texttt{FILE\_HEADER.Machine} fields with zero, a method previously employed by the SOREL dataset to prevent files from running \cite{sorel}. 
Furthermore, we also release EMBER raw features (feature version 2) for each malware sample \cite{ember}. In addition to being the largest public malware dataset with ground truth family labels as of the time of writing, MOTIF is the largest public dataset with ground truth family labels and a full alias mapping derived from both open-source reporting and antivirus signatures. 
The remainder of this section describes our methodology for processing the open-source threat intelligence reports, discusses how we resolved malware family aliases, and reviews the contents of the MOTIF dataset.

\subsection{Source and Report Inclusion}
\label{sec:inclusion}
Table \ref{tab:source_counts} lists the 14 sources that contributed to the MOTIF dataset and how many labeled malware hashes we identified from each. All sources are large, reputable cybersecurity organizations that regularly release threat reports containing IOCs to the community. We considered several other organizations that satisfy these requirements but due to time and resource constraints we were unable to review them. Thus, the selection of sources to include in the MOTIF dataset was somewhat subjective; the sources that were included tended to have the greatest name recognition and publish the most IOCs in their reports.

\begin{table}
\caption{MOTIF Sources}
\adjustbox{max width=\columnwidth}{%
\begin{tabular}{@{}lclc@{}}
\toprule
\multicolumn{1}{c}{Source} & \multicolumn{1}{r}{Samples} &  \multicolumn{1}{c}{Source} & \multicolumn{1}{r}{Samples} \\ \midrule
Bitdefender	\cite{Bitdefender} & 199 & G DATA \cite{G_Data} & 106 \\
CheckPoint \cite{Checkpoint} & 221 & Kaspersky \cite{Kaspersky} & 502 \\
CISA \cite{CISA} & 101 & Malwarebytes \cite{Malwarebytes} & 201 \\
Cybereason \cite{Cybereason} & 204 & Palo Alto Networks \cite{Palo_Alto_Networks} & 434 \\
ESET \cite{ESET} & 607 & Proofpoint \cite{Proofpoint} & 567 \\
FireEye \cite{FireEye} & 314 & Symantec \cite{Symantec} & 164 \\
Fortinet \cite{Fortinet} & 269 & Talos \cite{Talos} & 480 \\
\bottomrule
\end{tabular}
}
\label{tab:source_counts}
\end{table}

We reviewed \textit{all} open-source threat intelligence reports published by the 14 sources listed in Table \ref{tab:source_counts} between January 1, 2016 and January 1, 2021. This time window permits the inclusion of recent malware (as of the time of writing) and should allow for most antivirus signatures to have stabilized \cite{zhu}. In order to ensure that the malware samples included in the MOTIF dataset have family labels with ground truth confidence, we processed each report in a standardized manner. Reports were omitted if they did not meet all of the following conditions: First, the report must provide a detailed technical analysis of the malware, implying that some expert manual analysis was performed. Second, the report must include the MD5, SHA-1, or SHA-256 hashes of the analyzed malware samples and clearly indicate their corresponding families. Finally, the report must be fully open-source: reports requiring email registration or a subscription to a paid service were not included. Of the thousands of reports we surveyed, only 644 met these conditions and contained at least one labeled hash.

\subsection{Malware Sample Inclusion and Naming} We also stipulated which malware samples in a report could be included in the dataset. Malware samples without a clear, unambiguous family label were omitted. Samples labeled only using an antivirus signature (and not a family name) were also skipped. Malware samples not targeting the Windows, macOS, or Linux operating systems were excluded. Large dumps of 50 or more malware samples were not included, unless very detailed analysis was provided which suggested all files received manual attention. We emphasize that while constructing the MOTIF dataset, we made every reasonable effort to be methodical and judicious about which malware samples were included.

The following conventions were used when cataloguing malware family names from reports. Any files used to download, launch, or load a malware sample were not treated as belonging to that sample's family. However, plugins and modules used by modular malware were considered to have family membership. Legitimate tools abused by malicious actors (e.g. Metasploit, Cobaltstrike, etc.) were not treated as malware families.
We recorded 4,369 malware hashes with 595 distinct family names (normalized, without alias resolution) during this procedure. 
Family names were normalized by converting them to lowercase and removing all non-alphanumeric characters. We later use the same process for normalizing the names of threat actors and, for the remainder of this paper, we use normalized names when referring to a malware family or threat actor.
In order to ensure reproducibility, we also recorded the name of the source, the date the report was published, and the URL of the report. In some reports the IOCs were provided in a separate appendix; a second URL to the appendix is included in these cases. As described at the beginning of Section \ref{sec:motif}, non-PE files and files to which we could not gain access were not included in MOTIF. As a result, MOTIF contains 3,095 files from 454 malware families.

In manual review of the reports, we have found no occurrences of two reports disagreeing on a malware family label. When a report included a non-PE file as part of the family (e.g., a PDF used as the payload for the malware), we excluded it and retained only the valid Windows PE executables. Only reports that provided well documented manual analysis supporting its conclusion were included, giving us high confidence in all samples of the MOTIF dataset.

\subsection{Family Alias Resolution}
\label{sec:alias_resolution}

The MOTIF dataset provides a comprehensive alias mapping for each family, in addition to a brief description of that family's capabilities and which specific threat actor or campaign it is attributed to, if any.  
Descriptions are succinct and based on our subjective expertise. Most descriptions contain the category of malware to which the family belongs (e.g. ransomware) and a noteworthy fact about the malware (e.g. an interesting capability, the language it was written in, or the name of a related family). In cases when a threat actor has multiple names, the 2-3 most common ones were provided.

Our primary method for identifying family aliases was open-source reporting. Threat reports about a particular family often supply the aliases used by other organizations. We were thorough in our search for aliases and only considered reports which we considered to be published by reputable organizations. We observed that the boundary between two malware families being aliases and variants is often nebulous. To remain consistent, we considered two families to be variants when significant functionality was added to a malware family and subsequent reports referred to the new samples by a different name. In cases when malware was "rebranded" but no new functionality was added (e.g. babax / osno) we treated the two family names as aliases.
As a secondary source of alias naming, we investigated antivirus signatures that contained slight variations (e.g. agenttesla / agensla) or rearrangements (e.g. remcos / socmer) of family names. In these cases, it was often obvious when two families were aliases. However, antivirus signatures that indicated a generic detection (e.g. centerposgen) could refer to multiple families or threat actors (e.g. the ekans / snake ransomware and the turla / snake APT) or were too generic (e.g. padcrypt / crypt), were disregarded.

Our investigation of malware family aliases was supplemented by Malpedia \cite{malpedia}, which provides information about family aliases, behavior, and attribution. We independently confirmed these details using open-source reporting from reputable organizations. In a few cases, we identified family names which we suspected were aliases, but we were unable to confirm a relationship using publicly available information. We assess that our alias mapping is as comprehensive as possible given available knowledge and that the impact upon evaluation from the small number of suspected missing aliases is negligible. The alias mapping contains 968 total aliases for the 454 families in MOTIF, an average of 2.132 per family. The wannacry family has an astonishing 15 aliases and 25 families in the dataset have at least five aliases. As illustrated by these results, we strongly feel that inconsistent family naming is a significant issue in the malware analysis community and 
continued effort to study and correct this problem is warranted.

\subsection{Malware Demographics in MOTIF}
\label{sec:demographics}

\begin{figure}
\caption{MOTIF Family Size Distribution}
    \centering
    \includegraphics[width=\columnwidth,keepaspectratio]{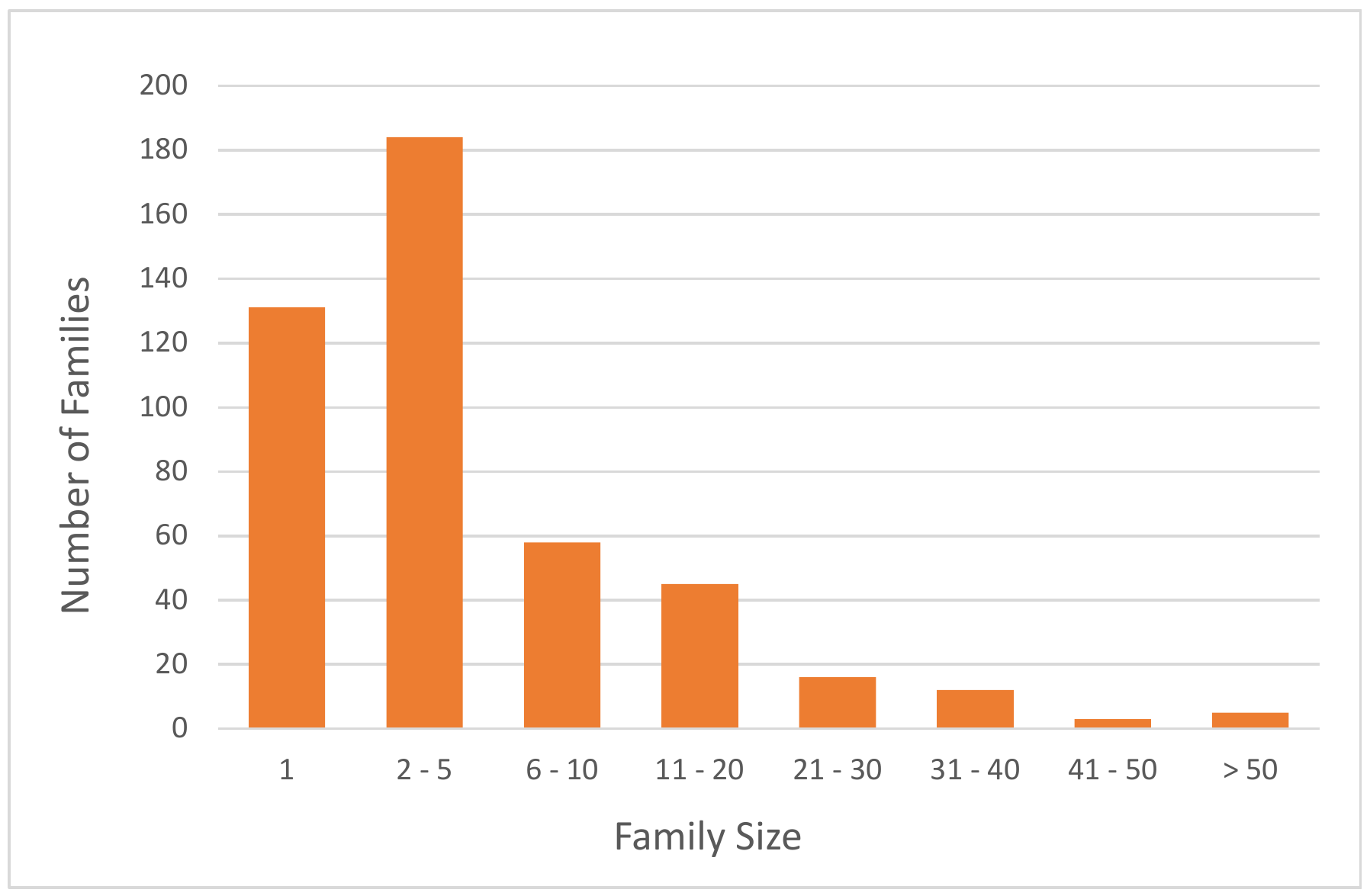}
    \label{fig:family_distr}
\end{figure}

Figure \ref{fig:family_distr} displays the distribution of malware family sizes in MOTIF. The vast majority of families in the dataset are represented by five or fewer samples, many by just one sample. The icedid banking trojan is by far the largest family in MOTIF, with 142 samples. Many of the sources we studied had clear tendencies as to which types of malware were included in their reports. For example, Fortinet \cite{Fortinet} focused primarily on malware associated with cybercrime, with a heavy emphasis on ransomware. CISA \cite{CISA} reported almost exclusively on state-sponsored malware, especially malware tied to North Korea. Many other sources also reported heavily on ransomware and malware used in targeted attacks, which is reflected in the composition of the MOTIF dataset. MOTIF contains 151 malware families attributed to 42 distinct threat actors or campaigns associated with targeted attacks. Criminal groups that do not engage in cyber-espionage (e.g. ta505 and the carbanak group) are not included in this tally.
Nearly a third of the malware samples in MOTIF (974 files) have been attributed to one of these 42 threat actors. Ransomware also makes up a significant portion of MOTIF, with 576 malware samples from 102 families. Backdoors, downloaders, RATs, infostealers, and ATM malware are also common malware categories. Adware and other types of PUAs were not frequently reported on and thus have little representation in MOTIF. Although MOTIF includes first-stage malware, such as droppers and downloaders, this category of malware was reported on far more often than is apparent in the dataset. This is likely because first-stage malware is  rarely granted a family designation unless it is associated with a targeted attack (e.g artfulpie) or is known for delivering notable commodity malware families (e.g. hancitor).

\subsection{Sources of Bias in MOTIF}

We acknowledge that the methods used to construct MOTIF biases the data. Due to the factors described in Section \ref{sec:demographics}, MOTIF does not reflect the average distribution of malware that might be encountered on a daily basis, but rather, it portrays a collection of malware that a cybersecurity organization would deem most \emph{important}. Furthermore, the malware samples in MOTIF were published in reports dating between January 1, 2016 and January 1, 2021. MOTIF is no exception to the rule that manually labeling malware is time-consuming, and the malware samples in MOTIF will become outdated over time. We consider MOTIF to be a ``hard" dataset on account of its high number of families and the considerable proportion of malware attributed to advanced threat actors. In addition, MOTIF contains multiple malware families that are variants of each other, are attributed to the same threat actor, are packed using the same packer, or share other similarities. We have high confidence that MOTIF is more challenging than existing datasets and we provide further evidence of this claim in Sections \ref{sec:avclass_eval} and \ref{sec:machinelearning}. At the same time, MOTIF represents a scant 3,095 malware samples in an much larger ecosystem where hundreds of thousands of new malware samples are being observed daily \cite{virustotal_stats}. Creating larger, even more representative datasets is a goal of future work.

\section{Experiments}
\label{sec:experiments}

To date, MOTIF is the largest public malware dataset with expert ground truth family labels. It is also one of the most diverse, with 454 distinct families. Unlike prior datasets, many of which have approximate labels, lack diversity, or contain outdated malware samples, we claim that evaluations made using the MOTIF dataset can be made with confidence. Furthermore, MOTIF's combination of ground truth confidence family labels and a comprehensive alias mapping enable experiments and evaluations that could not be performed before. Machine learning has been investigated for malware classification since 1995 \cite{Kephart:1995:BID:1625855.1625983}, and its study has used techniques from across the spectrum of classical ML methods like SVMs and boosting, deep learning, graph based learning, supervised and unsupervised clustering, and borrowing from Computer Vision and Natural Language Processing techniques is frequent ~\cite{Raff2020a}. It is not possible to enumerate all current relevant ML approaches. In this section, we evaluate a notable malware classifier and multiple clustering algorithms using the MOTIF dataset. In addition, we provide benchmark results for two ML models trained on MOTIF, and we assess the capabilities of three outlier detection models to identify novel malware families. The MOTIF dataset enables us to obtain new insights about these tasks and to identify new directions for future research. %
For this section we use Precision and Recall in the clustering specific terminology relating to the ability to group different families into different clusters and same families into same clusters respectively \cite{bitshred}. This style of defining Precision and Recall is prevalent in the malware literature in part due to the difficulty in knowing what the true class labels are, which MOTIF helps resolve. As our results will show, these alternate definitions of necessity can be misleading, ``hiding'' instances of data points that are being labeled incorrectly in a consistent fashion. 

\subsection{Evaluating AVClass}
\label{sec:avclass_eval}

In order for MOTIF to be a benchmark for evaluating antivirus-based malware classifiers, all samples have been uploaded to VirusTotal \cite{virustotal}, a platform that scans malware samples with a large number of antivirus products. In our first experiment we evaluate AVClass using the MOTIF dataset. AVClass processes and normalizes antivirus scan reports and then uses plurality voting to predict a malware family \cite{avclass}. AVClass is one of the most widely used approaches to better automated family labeling, and operates on the family labels produced by several Anti-Virus products. Using domain specific steps akin to tokenization, stop-word removal, and lemmatization to resolve inconsistencies between products. We obtained antivirus scan reports for each of the 3,095 malware samples in MOTIF by querying the VirusTotal API. All queries used in the following experiments were made in Aug. 2021. Although the VirusTotal terms of service prohibit us from distributing these scan reports, they can easily be obtained by querying the VirusTotal API.

Table \ref{tab:avclassmetrics} displays AVClass's precision, recall, and F1 measure on five public malware datasets when run under default settings, previously reported by \citet{avclass}. Note that Malgenome* represents a modification to the MalGenome dataset that groups six variants of DroidKungFu into a single family. Table \ref{tab:avclassmetrics} also shows AVClass's evaluation results on MOTIF when using its default alias mapping (MOTIF-Default) and when provided with MOTIF's alias mapping (MOTIF-Alias). In both cases, all of AVClass's evaluation metrics are significantly lower for MOTIF than any other 
\begin{table}[!h]
\caption{AVClass Evaluation Results}
\adjustbox{max width=\columnwidth}{%
\label{tab:avclassmetrics}
\begin{tabular}{@{}lrrrr@{}}
\toprule
Dataset &  Precision & Recall & F1 Measure & Accuracy\\ \midrule
    Drebin & 0.954 & 0.884 & 0.918 & Unknown\\
    Malicia & 0.949 & 0.680 & 0.792 & Unknown\\
    Malsign & 0.904 & 0.907 & 0.905 & Unknown\\
    MalGenome* & 0.879 & 0.933 & 0.926 & Unknown\\
    Malheur & 0.904 & 0.983 & 0.942 & Unknown\\ 
    MOTIF-Default & \textbf{0.763} & \textbf{0.674} & \textbf{0.716} & \textbf{0.468}\\
    MOTIF-Alias & 0.773 & 0.700 & 0.735 & 0.506\\
    \bottomrule
\end{tabular}
}
\end{table}
dataset, with the exception of recall for the Malicia dataset. This raises the question of why the evaluation results of the other datasets are so much higher. Drebin (the dataset with the highest precision) and Malheur (the dataset with the highest recall and F1 measure) were both labeled using antivirus majority voting. Because this is very similar to AVClass's plurality voting strategy, evaluation results on these datasets were likely artificially high. Other attributes of the other datasets used to evaluate AVClass, such as containing Android malware rather than Windows, having outdated malware samples, using approximate labeling methods, or lacking size or family diversity, may have also contributed to the observed discrepancies.

Because computing the accuracy of a malware classifier requires a mapping between the aliases used by the classifier and the aliases used by the reference dataset, the majority of works only use precision and recall, which are based on accurate grouping (and not necessarily accurate labeling) \cite{avclass,malsign,nappa,rieck}. The MOTIF dataset provides a full family alias mapping, allowing accuracy to be computed as well. In Table \ref{tab:avclassmetrics} the stark contrast between AVClass's accuracy on the MOTIF dataset (46.78\%) and its precision and recall (76.35\% and 67.36\% respectively)
is evident. Although our results show that AVClass labels related malware samples with some consistency, the tool predicts an incorrect family name for a malware sample more often than the correct one.

\subsection{Further Investigation of Antivirus Results}

An investigation of the labels predicted by AVClass in our prior experiment revealed that errors were caused in part by AVClass, but primarily due to the antivirus scan reports used as input. Antivirus signatures frequently contained the name of a variant of the correct family, the name of a family with similar capabilities to the correct one, or the name of a family known to be otherwise associated with the correct one. Furthermore, antivirus signatures commonly contained non-family information that AVClass did not properly discard, including the name of the group that the sample is attributed to, the broad category of malware it belongs to, the sample's behavioral attributes, the name of the packer that the sample was packed with, or the programming language that the sample was written in. The discrepancies between the results for MOTIF-Default and MOTIF-Alias in Table \ref{tab:avclassmetrics} also indicate that AVClass also often fails to resolve family aliases properly. In our study, these factors frequently caused AVClass to make incorrect predictions; in many cases, the predicted labels were not the name of any valid malware family. Precision and recall are intended to be used as cluster evaluation metrics and it is clear that model performance can be severely misjudged in cases such as these. Future malware datasets should continue to offer comprehensive alias mappings for each of their constituent families so that accuracy can be computed.
Further investigation revealed that of the 3,095 antivirus scans reports for the MOTIF dataset, 577 reports (18.64\%) do not include the correct family name at all and 934 reports (30.18\%) include it once at most. The lack of family information in these scan reports would prevent an antivirus-based malware family classifier from achieving higher than 81.36\% accuracy on the MOTIF dataset.

By running AVClass with MOTIF's family alias mapping, we extracted family information from each scan report for the MOTIF dataset. Rather than applying plurality voting (AVClass's voting method) to these results, we instead used majority voting.
Antivirus majority voting resulted in 1,178 reports with the correct family, 719 reports with an incorrect family, and 1,198 reports where no family was the clear majority. After discarding the scan reports with no majority (as is common practice), antivirus majority voting resulted in only 62.10\% accuracy. In Section \ref{sec:demographics} we noted that MOTIF contains a disproportionate amount of malware attributed to threat actors associated with targeted attacks. To test whether this was impacting our results, we repeated the experiment using only the 2,121 malware samples from MOTIF that are not attributed to one of these threat actors. Although slightly improved, the results were similar - 854 scan reports had no clear majority family, and only 863 of the remaining 1,267 reports (68.11\%) had a correct majority vote. Conventional wisdom has always held that antivirus majority voting is highly accurate, and no prior work has challenged this assumption \cite{avclass}. However, our finding that antivirus majority voting has just 62.10\% accuracy on MOTIF indicates that this belief may require re-examination.

\subsection{Evaluating Metadata Hashes}

Next, we evaluate the precision, recall, and F1 measure of four hashing algorithms used for identifying similar malware samples. These hashes use the metadata of the PE files, and are widely used as tools to identify similar files. To the best of our knowledge such metadata hashes have not been evaluated on a single corpus for comparison, and the design of new hashes using \textit{Learning to Hash} \cite{Wang2018,NIPS2009_3667} research is an open problem MOTIF enables. 
Two PE files have identical Imphash digests if their Import Address Tables (IATs) contain the same functions in the same order \cite{imphash}. If two files have equivalent values for specific metadata fields in their \texttt{FILE\_HEADER}, \texttt{OPTIONAL\_HEADER}, and each \texttt{SECTION\_HEADER}, they have the same peHash digest \cite{pehash}. Finally, the RichPE hash can be computed for PE files containing a Rich header, which is added to files compiled and linked using the Microsoft toolchain. Files that share select values in their Rich header, \texttt{FILE\_HEADER}, and \texttt{OPTIONAL\_HEADER} have the same RichPE digest \cite{richpe}. The remaining hash - vHash - is VirusTotal's ``simple structural feature hash'' \cite{vhash}. VirusTotal has not disclosed how the vHash of a file is computed, and we don't know whether it is based upon metadata or file contents. We are aware of no formal evaluation of vHash. Although prior evaluation has shown that the three remaining hashes are effective at grouping similar malware, we are unaware of any studies that quantify their precision or recall using a reference dataset with family labels.

\vspace*{-4pt}
\begin{table}[!h]
\caption{Metadata Hash Evaluation Results}
\centering
\adjustbox{max width=\columnwidth}{%
\label{tab:hash_results}
\begin{tabular}{@{}lrrr@{}}
\toprule
Hash Name & Precision & Recall & F1 Measure\\ \midrule
    Imphash & 0.866 & \textbf{0.382} & \textbf{0.530}\\
    Imphash-10* & 0.971 & 0.301 & 0.460\\
    peHash & \textbf{0.998} & 0.264 & 0.417\\
    RichPE & 0.964 & 0.331 & 0.494\\
    vHash & 0.983 & 0.317 & 0.480\\
    \bottomrule
\end{tabular}
}
\end{table}

In order to better understand how often collisions between unrelated malware samples occur and how effective these hashes are at grouping malware from the same family, we clustered the MOTIF dataset on identical hash digests. Files for which a hash digest could not be determined (e.g. PE files without an IAT or Rich header) were assigned to singleton clusters. Table \ref{tab:hash_results} displays evaluation results for each hash. Imphash has the highest recall and F1 measure, while at the same time having the lowest precision. Files with few imports may not have a unique Imphash \cite{imphash}, so we repeated the evaluation, assigning all files with fewer than 10 imports to singleton clusters. The results (denoted Imphash-10 in Table \ref{tab:hash_results}) indicate that this modification drastically increases the precision of Imphash, but causes a correspondingly large drop in recall and F1 measure. We were not surprised that peHash had a near-perfect precision, as it is widely regarded by the community as a very strict metadata hash. However, this property also yielded the lowest recall and F1 measure. Finally, although RichPE and vHash had no outstanding metric results compared to the other hashes, both possess high precision values. All recall results seem to be poor, but this is typical of metadata hashes as even small changes in a file's metadata can result in a different digest.

\subsection{Machine Learning Experiments}
\label{sec:machinelearning}

MOTIF provides the first rigorous benchmark for ML family classification methods. We demonstrate this using malware family classification and novel family detection, two standard ML tasks performed extensively on prior datasets \cite{kaggle,drebin,ember,sorel}. We trained LightGBM~\cite{NIPS2017_6907} and MalConv2~\cite{Raff2020b} models on the MOTIF dataset using default settings. The LightGBM model uses EMBER feature vectors, while MalConv2 was trained on disarmed binaries using a single Titan RTX GPU.
\begin{table}[!h]
\centering
\caption{Few-Shot Learning Results}
\adjustbox{max width=\columnwidth}{%
\label{tab:ml_results}
\begin{tabular}{@{}lrrr@{}}
\toprule
Model &  Accuracy & Std. Dev.\\ \midrule
LightGBM & \textbf{0.724} & 0.021\\
MalConv2 & 0.487 & 0.017\\
    \bottomrule
\end{tabular}
}
\end{table}
Since malware families with only one representative sample (which we call singleton families) cannot be represented in both the training and test sets, they were left out of the experiment. Then, we performed five-fold stratified cross-validation on the remaining 2,964 files (323 families). The mean and standard deviation of the accuracy scores obtained during cross-validation are listed in Table \ref{tab:ml_results}. Neither model demonstrated particularly high accuracy, but they are both significantly better than random guessing (4.79\%). Furthermore, given the high number of families in the dataset and limited number of samples per family, their performances are very reasonable.

\begin{table}[!h]
\caption{Novel Family Detection Results}
\resizebox{\columnwidth}{!}{%
\label{tab:outlier_results}
\begin{tabular}{@{}lrrr@{}}
\toprule
Model & Precision & Recall & F1 Measure\\ \midrule
Isolation Forest & 0.0 & 0.0 & 0.0\\
Local Outlier Factor & \textbf{0.265} & 0.206 & 0.232\\
One-Class SVM & 0.233 & \textbf{0.382} & \textbf{0.289}\\
    \bottomrule
\end{tabular}
}
\end{table}

Malware family classifiers are trained on a finite number of malware families and new, unknown families will be present when a model is deployed into a production environment. To test how well ML models can distinguish novel malware families from known families, we withheld the 131 malware samples from singleton families in MOTIF, in addition to 10\% of the remaining dataset (297 files). Three outlier detection models were trained on EMBER feature vectors for the remainder of the files. The objective of the experiment was to determine whether these models would detect the singleton families as outliers because they were not included in the training set. As shown in Table \ref{tab:outlier_results}, none of the models were able to perform this task reliably and the Isolation Forest did not detect a single file as an outlier. Although existing ML models can distinguish between malware families with moderate accuracy, the overall differences between existing and novel families seem difficult to identify. Further research is needed to ensure that models can appropriately address novel families.

\section{Discussion and Conclusion}
\label{sec:conclusion}

MOTIF is the first large, publicly accessible corpus of Windows malware with ground truth reference labels. The disarmed malware samples, EMBER features, and linked reports are valuable resources for future research. MOTIF is also the first dataset to provide a comprehensive mapping of malware family aliases, enabling numerous experiments and evaluations that could not be previously performed. Results obtained using the MOTIF dataset have already challenged conventional wisdom firmly held by the community, such as the accuracy of techniques which use collective decisions of a group of antivirus products as a source of family labeling. In the first evaluations of their kind, we found that AVClass has a 46.78\% accuracy on the MOTIF dataset that is considerably lower than previously thought, and antivirus majority voting correctly classifies only 62.10\% of the malware samples for which a clear majority could be obtained. These findings impact nearly all malware family classification research, especially related to antivirus-based labeling.

We recognize that the malware demographics in MOTIF do not reflect the distribution of malware families that might be encountered in the wild, and instead express the families which malware analysis organizations consider to be most relevant, a balance which has pros and cons. 
While collecting reports over a longer period of time or from more sources could further expand the corpus, it is unlikely to significantly change the current limitations of MOTIF.
The use of high-confidence methods for identifying related malware (g.g., peHash) could significantly increase the size of MOTIF, at the cost of losing full ground truth confidence. Our hope for future datasets is that they will be constructed with care to identify the trade-offs in scale, label quality, and diversity so that they can be used together to provide more accurate judgments. 

Beyond the standard tasks shown in this paper, MOTIF opens the door to many new avenues of ML research to malware problems. The reports can be used to explore few- and zero-shot learning to detect families before samples are available. The label quality allows exploring transfer learning from larger, less accurately labeled, corpora. The large number of families more representative of real-world diversity also allows more consideration to metrics and training approaches in the face of class-imbalanced learning \cite{JMLR:v18:16-365,10.1145/3338501.3357374,DELGAUDIO2014,Akbani:2004:ASV:3108498.3108507,He2009,Kubat1997,PratiIICAI09,Moskovitch2009a,Blagus2013,Prati2004,Japkowicz:2002:CIP:1293951.1293954}. We envision MOTIF becoming a valuable asset for evaluating malware family classifiers and for enabling future malware research.

\bibliographystyle{IEEEtranN}
\bibliography{sample-base}

\clearpage
\appendix
\section{Datasheets for Datasets}
\label{section:datasheets}
\begin{mdframed}[linecolor=\sectioncolor]
\section*{\textcolor{\sectioncolor}{
    MOTIVATION
}}
\end{mdframed}

    \textcolor{\sectioncolor}{\textbf{
    For what purpose was the dataset created?
    }
    Was there a specific task in mind? Was there
    a specific gap that needed to be filled? Please provide a description.
    } \\
    The MOTIF dataset was created because there is a dearth of malware family datasets with ground truth reference labels. It is meant to improve evaluation of malware family classifiers, especially those that use antivirus-based ones. \\
    
    \textcolor{\sectioncolor}{\textbf{
    Who created this dataset (e.g., which team, research group) and on behalf
    of which entity (e.g., company, institution, organization)?
    }
    } \\
    This dataset was created as a collaboration between Booz Allen Hamilton and the University of Marlyand, Baltimore County. \\
    
    \textcolor{\sectioncolor}{\textbf{
    What support was needed to make this dataset?
    }
    (e.g.who funded the creation of the dataset? If there is an associated
    grant, provide the name of the grantor and the grant name and number, or if
    it was supported by a company or government agency, give those details.)
    } \\
    Creation of this dataset was supported by Booz Allen Hamilton. \\
    
    \textcolor{\sectioncolor}{\textbf{
    Any other comments?
    }} \\
    N/A. \\

\begin{mdframed}[linecolor=\sectioncolor]
\section*{\textcolor{\sectioncolor}{
    COMPOSITION
}}
\end{mdframed}
    \textcolor{\sectioncolor}{\textbf{
    What do the instances that comprise the dataset represent (e.g., documents,
    photos, people, countries)?
    }
    Are there multiple types of instances (e.g., movies, users, and ratings;
    people and interactions between them; nodes and edges)? Please provide a
    description.
    } \\
    The main instances of the dataset are disarmed malware samples. Raw EMBER features (representing extracted PE metadata) are also provided for each instance.
     \\

    \textcolor{\sectioncolor}{\textbf{
    How many instances are there in total (of each type, if appropriate)?
    }
    } \\
    There are 3,095 malware samples in total from 454 families, and a distribution of family sizes is located in Section \ref{sec:demographics}.\\
    
    \textcolor{\sectioncolor}{\textbf{
    Does the dataset contain all possible instances or is it a sample (not
    necessarily random) of instances from a larger set?
    }
    If the dataset is a sample, then what is the larger set? Is the sample
    representative of the larger set (e.g., geographic coverage)? If so, please
    describe how this representativeness was validated/verified. If it is not
    representative of the larger set, please describe why not (e.g., to cover a
    more diverse range of instances, because instances were withheld or
    unavailable).
    } \\
    The dataset does not contain all known malware samples, or all known malware samples from open-source threat reports. We provided detailed discussion of representativeness in Section \ref{sec:demographics}. It represents a sampling of malware that cybersecurity organzations would find notable. \\

    \textcolor{\sectioncolor}{\textbf{
    What data does each instance consist of?
    }
    “Raw” data (e.g., unprocessed text or images) or features? In either case,
    please provide a description.
    } \\
    The disarmed malware samples are raw data (sequences of bytes) and the EMBER metadata are raw features. We also provide EMBERv2 feature vectors generated from the EMBER raw features. \\
    
    \textcolor{\sectioncolor}{\textbf{
    Is there a label or target associated with each instance?
    }
    If so, please provide a description.
    } \\
    Yes, each instance contains a malware family label. The name of the family, known aliases of the family, and a unique ID of the family that can be used as a ML label are provided.\\
    
    \textcolor{\sectioncolor}{\textbf{
    Is any information missing from individual instances?
    }
    If so, please provide a description, explaining why this information is
    missing (e.g., because it was unavailable). This does not include
    intentionally removed information, but might include, e.g., redacted text.
    } \\
    We assess that our alias mapping is comprehensive, but we suspect that a low number of aliases may be missing (not enough to impact the results discussed in our paper), either due to lack of reporting or lack of antivirus data. \\
    
    \textcolor{\sectioncolor}{\textbf{
    Are relationships between individual instances made explicit (e.g., users’
    movie ratings, social network links)?
    }
    If so, please describe how these relationships are made explicit.
    } \\
    Malware samples with the same family or attributed to the same threat group or campaign are  noted. We often note when two families are variants in the brief descriptions, but this information is likely not complete. \\
    
    \textcolor{\sectioncolor}{\textbf{
    Are there recommended data splits (e.g., training, development/validation,
    testing)?
    }
    If so, please provide a description of these splits, explaining the
    rationale behind them.
    } \\
    N/A. \\
    
    \textcolor{\sectioncolor}{\textbf{
    Are there any errors, sources of noise, or redundancies in the dataset?
    }
    If so, please provide a description.
    } \\
    Labels in this dataset are considered to have ground truth confidence. Any errors would be caused by an analyst writing a report or by the author who aggregated the dataset. \\
    
    \textcolor{\sectioncolor}{\textbf{
    Is the dataset self-contained, or does it link to or otherwise rely on
    external resources (e.g., websites, tweets, other datasets)?
    }
    If it links to or relies on external resources, a) are there guarantees
    that they will exist, and remain constant, over time; b) are there official
    archival versions of the complete dataset (i.e., including the external
    resources as they existed at the time the dataset was created); c) are
    there any restrictions (e.g., licenses, fees) associated with any of the
    external resources that might apply to a future user? Please provide
    descriptions of all external resources and any restrictions associated with
    them, as well as links or other access points, as appropriate.
    } \\
    This dataset links to fourteen external sources of open-source malware reporting (described in Table \ref{tab:source_counts}, with links provided in the References.). There are no guarantees that this data will continue to exist and remain constant, and many sources do not have official archives.  \\
    
    \textcolor{\sectioncolor}{\textbf{
    Does the dataset contain data that might be considered confidential (e.g.,
    data that is protected by legal privilege or by doctor-patient
    confidentiality, data that includes the content of individuals’ non-public
    communications)?
    }
    If so, please provide a description.
    } \\
    No. \\
    
    \textcolor{\sectioncolor}{\textbf{
    Does the dataset contain data that, if viewed directly, might be offensive,
    insulting, threatening, or might otherwise cause anxiety?
    }
    If so, please describe why.
    } \\
    No. \\
    
    \textcolor{\sectioncolor}{\textbf{
    Does the dataset relate to people?
    }
    If not, you may skip the remaining questions in this section.
    } \\
    No. \\
    
    \textcolor{\sectioncolor}{\textbf{
    Does the dataset identify any subpopulations (e.g., by age, gender)?
    }
    If so, please describe how these subpopulations are identified and
    provide a description of their respective distributions within the dataset.
    } \\
    N/A. \\

    \textcolor{\sectioncolor}{\textbf{
    Is it possible to identify individuals (i.e., one or more natural persons),
    either directly or indirectly (i.e., in combination with other data) from
    the dataset?
    }
    If so, please describe how.
    } \\
    N/A. \\
    
    \textcolor{\sectioncolor}{\textbf{
    Does the dataset contain data that might be considered sensitive in any way
    (e.g., data that reveals racial or ethnic origins, sexual orientations,
    religious beliefs, political opinions or union memberships, or locations;
    financial or health data; biometric or genetic data; forms of government
    identification, such as social security numbers; criminal history)?
    }
    If so, please provide a description.
    } \\
    N/A. \\
    
    \textcolor{\sectioncolor}{\textbf{
    Any other comments?
    }} \\
    N/A. \\

\begin{mdframed}[linecolor=\sectioncolor]
\section*{\textcolor{\sectioncolor}{
    COLLECTION
}}
\end{mdframed}

    \textcolor{\sectioncolor}{\textbf{
    How was the data associated with each instance acquired?
    }
    Was the data directly observable (e.g., raw text, movie ratings),
    reported by subjects (e.g., survey responses), or indirectly
    inferred/derived from other data (e.g., part-of-speech tags, model-based
    guesses for age or language)? If data was reported by subjects or
    indirectly inferred/derived from other data, was the data
    validated/verified? If so, please describe how.
    } \\
    We provide a full and detailed description of the process used to gather the data in Section \ref{sec:motif}.  \\
    
    \textcolor{\sectioncolor}{\textbf{
    Over what timeframe was the data collected?
    }
    Does this timeframe match the creation timeframe of the data associated
    with the instances (e.g., recent crawl of old news articles)? If not,
    please describe the timeframe in which the data associated with the
    instances was created. Finally, list when the dataset was first published.
    } \\
    Data was collected between Feb. and Aug. 2021. The surveyed articles were published between Jan. 2016 and Jan. 2021.  \\
    
    \textcolor{\sectioncolor}{\textbf{
    What mechanisms or procedures were used to collect the data (e.g., hardware
    apparatus or sensor, manual human curation, software program, software
    API)?
    }
    How were these mechanisms or procedures validated?
    } \\
    All data was collected manually from open-source threat intelligence reports. We validated the MOIF dataset in the following ways. We confirmed that there were no instances of reports disagreeing about the family for a malware sample. Additionally, we confirmed that all files matched the expected file type. We did find a small number of author errors using this method, in which the hash of a first-stage malware sample described in the report (such as a malicious document) was reported with the IOCs of the payloads for a family (usually executable files). These instances were removed from the MOTIF dataset. We will update the manuscript to reflect how we validated the results. \\
    
    \textcolor{\sectioncolor}{\textbf{
    What was the resource cost of collecting the data?
    }
    (e.g. what were the required computational resources, and the associated
    financial costs, and energy consumption - estimate the carbon footprint.
    See Strubell \textit{et al.}\cite{DBLP:journals/corr/abs-1906-02243} for approaches in this area.)
    } \\
    The financial costs amounted to compensating one data scientist during the time of data collection. No significant compute was needed for its construction beyond a standard laptop.\\ 
    
    \textcolor{\sectioncolor}{\textbf{
    If the dataset is a sample from a larger set, what was the sampling
    strategy (e.g., deterministic, probabilistic with specific sampling
    probabilities)?
    }
    } \\
    N/A. \\
    
    \textcolor{\sectioncolor}{\textbf{
    Who was involved in the data collection process (e.g., students,
    crowdworkers, contractors) and how were they compensated (e.g., how much
    were crowdworkers paid)?
    }
    } \\
    
    One Booz Allen Hamilton employee was involved in the data collection process and they were compensated per their normal salary. \\
    
    \textcolor{\sectioncolor}{\textbf{
    Were any ethical review processes conducted (e.g., by an institutional
    review board)?
    }
    If so, please provide a description of these review processes, including
    the outcomes, as well as a link or other access point to any supporting
    documentation.
    } \\
    No ethical review process was implemented. \\
    
    \textcolor{\sectioncolor}{\textbf{
    Does the dataset relate to people?
    }
    If not, you may skip the remainder of the questions in this section.
    } \\
    N/A. \\
    
    \textcolor{\sectioncolor}{\textbf{
    Did you collect the data from the individuals in question directly, or
    obtain it via third parties or other sources (e.g., websites)?
    }
    } \\
    N/A. \\
    
    \textcolor{\sectioncolor}{\textbf{
    Were the individuals in question notified about the data collection?
    }
    If so, please describe (or show with screenshots or other information) how
    notice was provided, and provide a link or other access point to, or
    otherwise reproduce, the exact language of the notification itself.
    } \\
    N/A. \\
    
    \textcolor{\sectioncolor}{\textbf{
    Did the individuals in question consent to the collection and use of their
    data?
    }
    If so, please describe (or show with screenshots or other information) how
    consent was requested and provided, and provide a link or other access
    point to, or otherwise reproduce, the exact language to which the
    individuals consented.
    } \\
    N/A. \\
    
    \textcolor{\sectioncolor}{\textbf{
    If consent was obtained, were the consenting individuals provided with a
    mechanism to revoke their consent in the future or for certain uses?
    }
     If so, please provide a description, as well as a link or other access
     point to the mechanism (if appropriate)
    } \\
    N/A. \\
    
    \textcolor{\sectioncolor}{\textbf{
    Has an analysis of the potential impact of the dataset and its use on data
    subjects (e.g., a data protection impact analysis) been conducted?
    }
    If so, please provide a description of this analysis, including the
    outcomes, as well as a link or other access point to any supporting
    documentation.
    } \\
    N/A. \\
    
    \textcolor{\sectioncolor}{\textbf{
    Any other comments?
    }} \\
    N/A. \\

\begin{mdframed}[linecolor=\sectioncolor]
\section*{\textcolor{\sectioncolor}{
    PREPROCESSING / CLEANING / LABELING
}}
\end{mdframed}

    \textcolor{\sectioncolor}{\textbf{
    Was any preprocessing/cleaning/labeling of the data
    done(e.g.,discretization or bucketing, tokenization, part-of-speech
    tagging, SIFT feature extraction, removal of instances, processing of
    missing values)?
    }
    If so, please provide a description. If not, you may skip the remainder of
    the questions in this section.
    } \\
    Malware hashes for which we did not have access to the corresponding files were not included in the MOTIF dataset.  Malware family names were normalized by converting them to lowercase and removing non-alphanumeric characters. \\

    \textcolor{\sectioncolor}{\textbf{
    Was the “raw” data saved in addition to the preprocessed/cleaned/labeled
    data (e.g., to support unanticipated future uses)?
    }
    If so, please provide a link or other access point to the “raw” data.
    } \\
    Survey information from hashes with no corresponding files is located in motif\_reports.csv. The original names of normalized families can be determined easily by navigating to the provided report URLs. \\

    \textcolor{\sectioncolor}{\textbf{
    Is the software used to preprocess/clean/label the instances available?
    }
    If so, please provide a link or other access point.
    } \\
    No software was used for preprocessing. \\

    \textcolor{\sectioncolor}{\textbf{
    Any other comments?
    }} \\
    N/A \\

\begin{mdframed}[linecolor=\sectioncolor]
\section*{\textcolor{\sectioncolor}{
    USES
}}
\end{mdframed}

    \textcolor{\sectioncolor}{\textbf{
    Has the dataset been used for any tasks already?
    }}

        So far, the only tasks this dataset has been used for are the experiments described in Section \ref{sec:experiments}. \\

    \textcolor{\sectioncolor}{\textbf{
    Is there a repository that links to any or all papers or systems that use the dataset?
    }
    If so, please provide a link or other access point.
    } \\
    Yes. A GitHub repository containing the dataset and code for training the discussed machine learning models is located at
    \hyperlink{https://github.com/boozallen/MOTIF}{https://github.com/boozallen/MOTIF}. \\

    \textcolor{\sectioncolor}{\textbf{
    What (other) tasks could the dataset be used for?
    }
    } \\
    This dataset can be used for evaluation of malwere family classifiers, ML and NLP tasks involving malware samples and reports written about them, and a variety of other malware clustering and classification tasks.\\

    \textcolor{\sectioncolor}{\textbf{
    Is there anything about the composition of the dataset or the way it was
    collected and preprocessed/cleaned/labeled that might impact future uses?
    }
    For example, is there anything that a future user might need to know to
    avoid uses that could result in unfair treatment of individuals or groups
    (e.g., stereotyping, quality of service issues) or other undesirable harms
    (e.g., financial harms, legal risks) If so, please provide a description.
    Is there anything a future user could do to mitigate these undesirable
    harms?
    } \\
    No. \\

    \textcolor{\sectioncolor}{\textbf{
    Are there tasks for which the dataset should not be used?
    }
    If so, please provide a description.
    } \\
    Use of this dataset must follow the terms of licensing at \hyperlink{https://github.com/boozallen/Public-License/blob/master/LICENSE.md}{https://github.com/boozallen/Public-License/blob/master/LICENSE.md} \\    %

    \textcolor{\sectioncolor}{\textbf{
    Any other comments?
    }}
    N/A. \\

\begin{mdframed}[linecolor=\sectioncolor]
\section*{\textcolor{\sectioncolor}{
    DISTRIBUTION
}}
\end{mdframed}

    \textcolor{\sectioncolor}{\textbf{
    Will the dataset be distributed to third parties outside of the entity
    (e.g., company, institution, organization) on behalf of which the dataset
    was created?
    }
    If so, please provide a description.
    } \\
    The dataset has been made fully open-source. \\

    \textcolor{\sectioncolor}{\textbf{
    How will the dataset will be distributed (e.g., tarball on website, API,
    GitHub)?
    }
    Does the dataset have a digital object identifier (DOI)?
    } \\
    The dataset is available on GitHub at \hyperlink{https://github.com/boozallen/MOTIF}{https://github.com/boozallen/MOTIF}. \\

    \textcolor{\sectioncolor}{\textbf{
    When will the dataset be distributed?
    }
    } \\
    The dataset will be made public on 12/1/2021. \\

    \textcolor{\sectioncolor}{\textbf{
    Will the dataset be distributed under a copyright or other intellectual
    property (IP) license, and/or under applicable terms of use (ToU)?
    }
    If so, please describe this license and/or ToU, and provide a link or other
    access point to, or otherwise reproduce, any relevant licensing terms or
    ToU, as well as any fees associated with these restrictions.
    } \\
    
    The dataset and code will be distributed under the \hyperlink{https://github.com/boozallen/Public-License/blob/master/LICENSE.md}{Booz Allen Public License} which allows for use, modification, and public distribution by non-profits, academics, and commercial entities, but does not allow selling of the dataset or derivatives. The license also limits liability, a requirement given the malware contained in the corpus. \\

    \textcolor{\sectioncolor}{\textbf{
    Have any third parties imposed IP-based or other restrictions on the data
    associated with the instances?
    }
    If so, please describe these restrictions, and provide a link or other
    access point to, or otherwise reproduce, any relevant licensing terms, as
    well as any fees associated with these restrictions.
    } \\
    The reports used are subject to copyright by their original owners, and links to them are provided to avoid any copyright issues. The VirusTotal reports can be obtained by others using a free account, but we are prohibited in redistributing them ourselves by the VirusTotal license. No other restrictions on this data have been imposed by third parties. \\

    \textcolor{\sectioncolor}{\textbf{
    Do any export controls or other regulatory restrictions apply to the
    dataset or to individual instances?
    }
    If so, please describe these restrictions, and provide a link or other
    access point to, or otherwise reproduce, any supporting documentation.\\
    } \\
    U.S. Export control laws may apply to any work produced in the U.S., and are the responsibility of external parties to confirm if a license is needed. We have not imposed any export control ourselves and are not aware of any special regularization that this dataset may fall under. We do not guarantee or warranty that some export control requirement now or in the future may apply to this dataset.\\ 

    \textcolor{\sectioncolor}{\textbf{
    Any other comments?
    }} \\
    N/A. \\

\begin{mdframed}[linecolor=\sectioncolor]
\section*{\textcolor{\sectioncolor}{
    MAINTENANCE
}}
\end{mdframed}

    \textcolor{\sectioncolor}{\textbf{
    Who is supporting/hosting/maintaining the dataset?
    }
    } \\
    Booz Allen Hamilton \\

    \textcolor{\sectioncolor}{\textbf{
    How can the owner/curator/manager of the dataset be contacted (e.g., email
    address)?
    }
    } \\
    The lead curator can be contacted at joyce\_robert2@bah.com or joyce@umbc.edu. \\

    \textcolor{\sectioncolor}{\textbf{
    Is there an erratum?
    }
    If so, please provide a link or other access point.
    } \\
    No. \\

    \textcolor{\sectioncolor}{\textbf{
    Will the dataset be updated (e.g., to correct labeling errors, add new
    instances, delete instances)?
    }
    If so, please describe how often, by whom, and how updates will be
    communicated to users (e.g., mailing list, GitHub)?
    } \\
    We may perform very infrequent updates, if at all. Updates would be solely to correct any errors made during curation or in code. Updates will be communicated via GitHub. 
    \\

    \textcolor{\sectioncolor}{\textbf{
    If the dataset relates to people, are there applicable limits on the
    retention of the data associated with the instances (e.g., were individuals
    in question told that their data would be retained for a fixed period of
    time and then deleted)?
    }
    If so, please describe these limits and explain how they will be enforced.
    } \\
    N/A. \\

    \textcolor{\sectioncolor}{\textbf{
    Will older versions of the dataset continue to be
    supported/hosted/maintained?
    }
    If so, please describe how. If not, please describe how its obsolescence
    will be communicated to users.
    } \\
    Yes, prior versions of the dataset will be made available via GitHub, and obsolescence will be communicated via GitHub. \\

    \textcolor{\sectioncolor}{\textbf{
    If others want to extend/augment/build on/contribute to the dataset, is
    there a mechanism for them to do so?
    }
    If so, please provide a description. Will these contributions be
    validated/verified? If so, please describe how. If not, why not? Is there a
    process for communicating/distributing these contributions to other users?
    If so, please provide a description.
    } \\
    The dataset is fully open-source and other users are free to augment it. These contributions will not be validated or verified. \\

    \textcolor{\sectioncolor}{\textbf{
    Any other comments?
    }} \\
    N/A. \\

\end{document}